\documentclass[twoside]{article}

\usepackage[accepted]{aistats2024halarxiv}

\usepackage[round]{natbib}

\usepackage{bm}
\usepackage{enumitem}
\usepackage{hyperref}
\usepackage{cleveref}
\usepackage{xcolor}
\usepackage{graphicx}
\usepackage[outdir=./]{epstopdf}
\usepackage{caption}
\usepackage{tabularx}
\usepackage{amsfonts}
\usepackage{amsmath}
\usepackage{amssymb}
\usepackage{mathabx}
\usepackage{stfloats}
\usepackage{titlesec}
\usepackage{amsthm}
\usepackage{booktabs}

\newtheorem{definition}{Definition}
\newtheorem{property}{Property}

\DeclareMathOperator*{\argmax}{argmax}

\newcommand{\bs}{\boldsymbol}

\begin{document}

%

%

\twocolumn[

\aistatstitle{Gaussian process regression with Sliced Wasserstein Weisfeiler-Lehman graph kernels }

\aistatsauthor{Raphaël Carpintero Perez\textsuperscript{1,2} \: \: \: Sébastien Da Veiga\textsuperscript{3} \: \: \: Josselin Garnier\textsuperscript{2} \: \: \: Brian Staber \textsuperscript{1}}
\aistatsaddress{ \textsuperscript{1} Safran Tech, Digital Sciences \& Technologies, 78114 Magny-Les-Hameaux, France\\ \textsuperscript{2} Centre de Mathématiques Appliquées, Ecole Polytechnique, Institut Polytechnique de Paris, 91120 Palaiseau, France\\
\textsuperscript{3} Univ Rennes, Ensai, CNRS, CREST - UMR 9194, F-35000 Rennes, France}
]

\begin{abstract}
Supervised learning has recently garnered significant attention in the field of computational physics due to its ability to effectively extract complex patterns for tasks like solving partial differential equations, or predicting material properties. Traditionally, such datasets consist of inputs given as meshes with a large number of nodes representing the problem geometry (seen as graphs), and corresponding outputs obtained with a numerical solver. This means the supervised learning model must be able to handle large and sparse graphs with continuous node attributes. In this work, we focus on Gaussian process regression, for which we introduce the Sliced Wasserstein Weisfeiler-Lehman (SWWL) graph kernel. In contrast to existing graph kernels, the proposed SWWL kernel enjoys positive definiteness and a drastic complexity reduction, which  makes it possible to process datasets that were previously impossible to handle. The new kernel is first validated on graph classification for molecular datasets, where the input graphs have a few tens of nodes. The efficiency of the SWWL kernel is then illustrated on graph regression in computational fluid dynamics and solid mechanics, where the input graphs are made up of tens of thousands of nodes.


\end{abstract}

\section{INTRODUCTION}
Over the past decade, there has been a growing interest for efficient machine learning algorithms applied to graph data. Graph learning tasks have historically emerged in fields such as biochemistry and social recommendation systems, but very recently, learning physics-based simulations described by partial differential equations (PDEs) has attracted a lot of attention. Indeed, traditional methods such as finite element provide accurate approximations of the PDE solution in computational fluid and solids mechanics, but such mesh-based methods are known for being computationally expensive, and thus difficult to apply to real-world design problems. To circumvent this issue, many efforts have been dedicated to the development of neural networks in order to emulate solutions to physical systems, either by incorporating physical knowledge \citep{karniadakis2021physics}, or by designing efficient graph neural networks architectures \citep{pfaff2020}.
In this setting, the supervised learning task involves inputs given as graphs and outputs being physical quantities of interest. We focus here on kernel methods, and more specifically Gaussian process regression, since they are powerful when the sample size is small, such as in computational physics, and when uncertainty quantification is needed. But in most applications the input graphs correspond to finite element meshes with several tens of thousands of nodes and continuous attributes. There is a wealth of literature about kernels between graphs, but few approaches handle such graphs with continuous node attributes, and most of them are intractable with \emph{large} and \emph{sparse} graphs.

\begin{figure*}[ht]
\begin{center}
\includegraphics[scale=0.42]{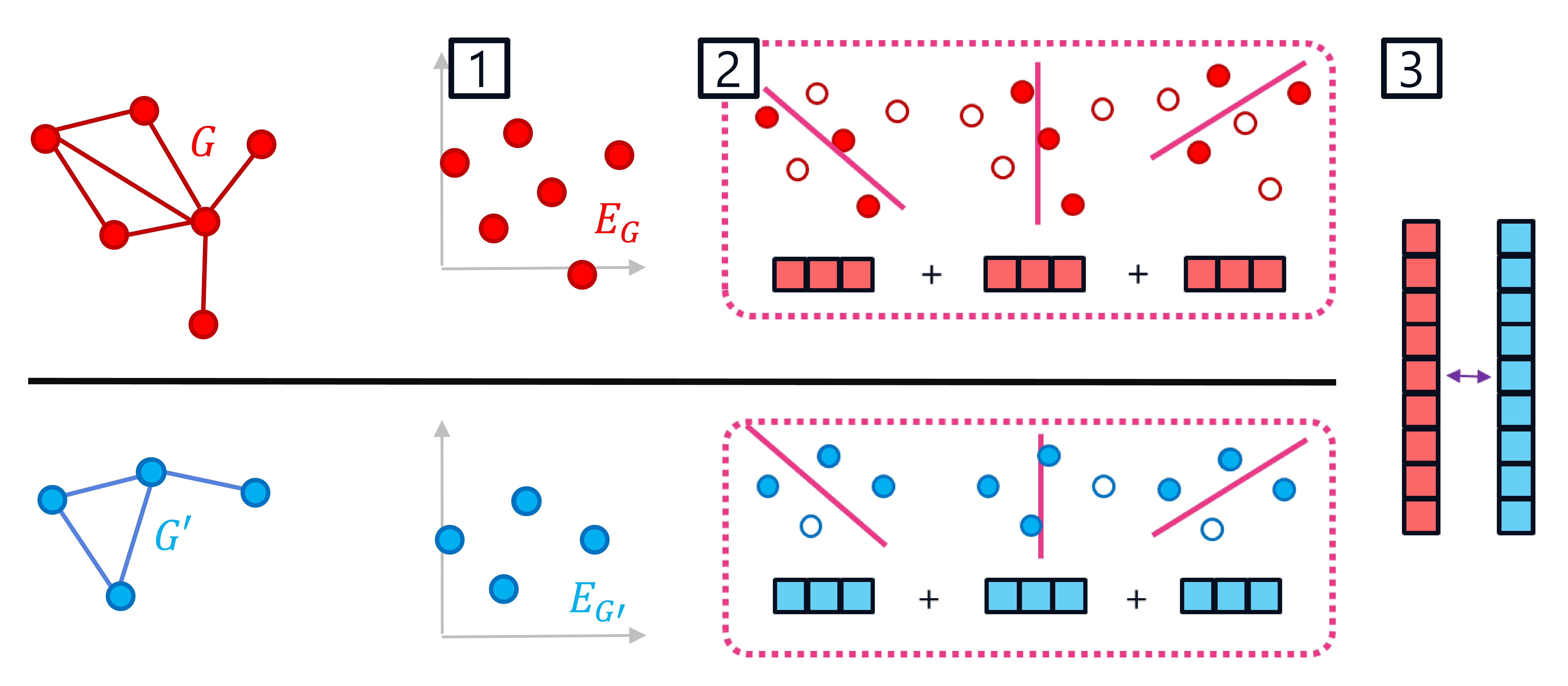}
\end{center}
\caption{SWWL kernel. Step 1: Graph embedding. Step 2: Projected quantile embeddings with only a few quantiles kept. Step 3: Euclidean distances between embeddings.}
\label{fig:swwl}
\end{figure*}

Herein, we propose the Sliced Wasserstein Weisfeiler-Lehman (SWWL) kernel as a variant of the Wasserstein Weisfeiler-Lehman (WWL) graph kernel proposed by \citet{wwl}. In a first step, unsupervised node embeddings are built using a continuous Weisfeiler-Lehman scheme. 
Then, instead of relying on the Wasserstein distance, the sliced Wasserstein distance between the empirical probability distributions of the augmented node attributes is computed. Figure \ref{fig:swwl} summarizes the proposed methodology. In addition to the drastic complexity reduction, the sliced Wasserstein (SW) distance is Hilbertian, thus allowing the construction of a positive definite kernel, which is not the case with the Wasserstein distance. Similarly to the WWL kernel, our approach creates unsupervised and non-parametric embeddings. The learning process can thus be broken down into two parts: preprocessing the input graphs, and assembling the Gram matrix with the optimisation of its hyperparameters. This is an advantage over related methods that have features built in a supervised manner. The estimation of SW distances involves random projections and empirical quantiles, and we show experimentally that good accuracies can be obtained with few projections and quantiles, thus resulting in a significant dimension reduction of the graph embedding without degrading performance.

Our contribution is three-fold:
\begin{itemize}[topsep=0pt]
    \itemsep0em 
    \item We design a new graph kernel called the Sliced Wasserstein Weisfeiler-Lehman kernel, which can handle large-scale graphs.
    \item We show the positive definiteness of this kernel.
    \item We demonstrate its efficiency when used in Gaussian process regression on high-dimensional simulation datasets from computational physics.
\end{itemize}
The article is organized as follows. Related work on graph kernels are discussed in Section \ref{section:related}. Section \ref{sec:preliminaries} first introduces the problem setting and recalls Gaussian process regression. The proposed SWWL kernel and its properties are described in Section \ref{section:swwl_for_gp}. Numerical experiments are then performed in Section \ref{section:exp}. The efficiency of the SWWL kernel is assessed first for classification tasks with molecular data, and for regression tasks in the case of mesh-based simulations.

\section{RELATED WORK} \label{section:related}


The problem of assessing similarity between graphs dates back to the 2000s, and a multitude of approaches have been proposed since then \citep{a_survey_on_gk}.
Among the rich literature about kernels on graphs \citep{gk_a_survey, gk_state_of_the_art}, it can be noted that most are not adapted to the case where graphs have continuous node attributes. This seriously hinders their applicability to computational physics, characterized by real-valued physical properties at the mesh nodes.

Complexity of such methods is another limitation: originally designed for molecular datasets where graphs have less than 100 nodes, such kernels do not scale up well to larger graphs. Last but not least, graph kernels often focus on graph structure by matching sub-patterns with the R-convolution framework \citep{R_convolution} such as shortest paths, random walks, graphlets or trees. However, this turns out to be irrelevant when the adjacency varies little between several inputs. The approach of \cite{hashing} extends some of the previous kernels to the continuous setting using multiple hashing functions.

Recently, many approaches focused on solving the graph-alignment problem using optimal transport and trying to find a matching between well-chosen distributions. The Fused-Gromov distance proposed by \cite{fgw} makes it possible to separate the information provided by the graph structure and its attributes with a trade-off between Wasserstein and Gromov-Wasserstein distances.
%
The recent approach of \cite{a_simple_way_to_learn_metrics_btw_attributed_graphs} uses embeddings obtained with Simple Graph Convolutional Network (SGCN) \citep{sgcn}, a simplified version of GCN reducing the number of trainable parameters.  
They also replace the Wasserstein distance by the Restricted Projected Wasserstein that uses projections to determine the optimal transport plans and their costs.
Other approaches propose to use the Linear Optimal Transport framework by computing transport displacements to a reference template (\cite{wegl}) or to regularize the Wasserstein distance using both local and global structural information of the graph (\cite{rwk}). 
However, all approaches using Wasserstein or Gromov-Wasserstein distances suffer from scalability issues and from the impossibility to obtain a positive definite kernel.

Graphs aside, many approaches tackle the problem of distribution regression, where a function over spaces of
probabilities is learnt \citep{smola2007hilbert, poczos2013distribution}. There are some positive definite kernels using optimal transport with possible entropic regularization \cite{bachoc2020gaussian, bachoc2023gaussian} but they require the use of a reference measure. Alternatively, the sliced Wasserstein distance can be used to define positive definite distance substitution kernels. It was initially proposed by \cite{kolouri2016} in order to build kernel functions between absolutely continuous distributions with respect to the Lebesgue measure and was extented to empirical distributions by \cite{meunier_sliced_pd}. Our work uses the results of \cite{meunier_sliced_pd} on the SW substitution kernel but differs in the SW approximation.

Finally, note that the notion of kernels between graphs is sometimes extended to fingerprint kernels for molecular data using for instance the Tanimoto/Jaccard kernel \citep{griffiths2022gauche}. However, such molecular kernels can only be used for this type of data and require a specific vector representation of the graphs, which is usually a complex problem.

\section{PRELIMINARIES} \label{sec:preliminaries}
We consider the task of learning a function $f : \mathcal{G} \rightarrow \mathcal{Y}$ where $\mathcal{Y} = \{0,1\}$ for classification tasks, and $\mathcal{Y} = \mathbb{R}$ for regression tasks. Here, $\mathcal{G}$ denotes a set of undirected and possibly weighted graphs having continuous attributes. Each graph $G \in \mathcal{G}$ can thus be represented as $G = (V, E, w, \mathbf{F})$ where $V$ is the set of 
$|V|$ nodes, and $E$ 
is a set of paired nodes, whose elements are called edges. The $d$-dimensional attributes of the nodes are gathered in the $(|V| \times d)$ matrix 
$\mathbf{F} = (\mathbf{F}_u)_{u \in V}$. 
The edge weights are assigned by the function $w : E \rightarrow \mathbb{R}$.
The neighborhood of a node $u \in V$ is given by $\mathcal{N}(u) = \{ v \in V: \{u,v\} \in E \}$ and its degree is denoted by $\mathrm{deg}(u)=|\mathcal{N}(u)|$.

We assume that we are given a dataset $\mathcal{D}$ consisting of $N$ observations $\mathcal{D} = \{ (G_i, y_i) \}^{N}_{i=1}$, where the input graphs may differ in terms of numbers of nodes and adjacency matrices, \textit{i.e.}, we may have $|V_i|\neq |V_j|$ and/or $E_i \neq E_j$ for some $(i, j) \in \{1, \ldots, N\}^2$. We consider the task of approximating function $f$ with kernel methods and in particular Gaussian process regression.

\begin{figure}[h]
\centering
\includegraphics[scale=0.28]{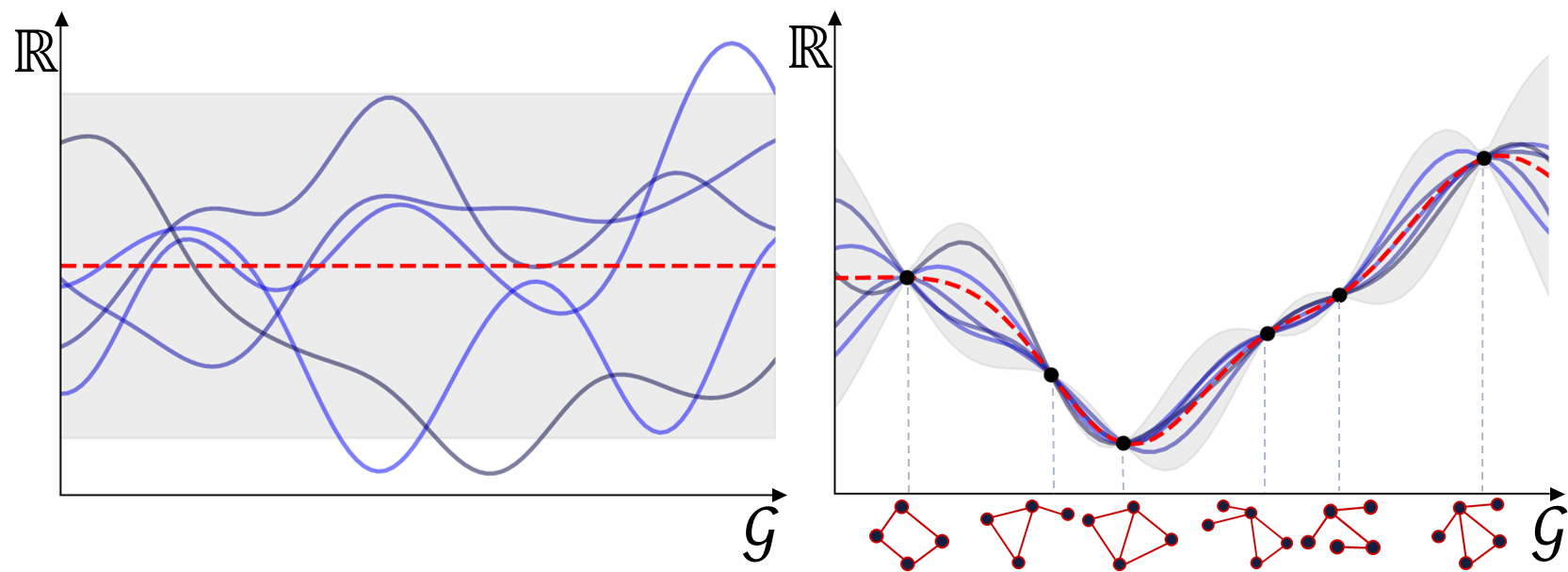}
\caption{Illustration of Gaussian process for graph inputs. Left: samples from the prior distribution. Right: samples from the posterior distribution after conditioning on observations (input points are graphs here).}
\label{fig:gp}
\end{figure}

\paragraph{Gaussian process regression.}
Gaussian process (GP) regression is a popular Bayesian approach for supervised learning in engineering \citep{gpml,gramacy2020surrogates,rgasp}. We assume that we are given noisy training observations  $y_i = f(G_i) + \epsilon_i$ for $i \in \{1, \ldots, N\}$ of $f:{\mathcal G}\to \mathbb{R}$ at input locations $\mathbf{G} = (G_i)_{i=1}^N$, where $\epsilon_i \sim \mathcal{N}(0,\eta^2)$ is a Gaussian additive noise. Let $\mathbf{f_*}:=(f({G}^*_i))_{i=1}^{N^*}$ be the values of $f$ at new test locations $\mathbf{G}^*=({G}^*_i)_{i=1}^{N^*}$. 
A zero-mean GP prior is placed on the function $f$. It follows that the joint distribution of the observed target values and the function values at the test locations writes (see, \textit{e.g.}, \citet{gpml})
\begin{equation}
    \begin{bmatrix}
    \mathbf{y}\\
    \mathbf{f}_*
    \end{bmatrix}
    \sim \mathcal{N} 
    \left( 
    \mathbf{0}, 
    \begin{bmatrix}
    \mathbf{K} + \eta^2 \mathbf{I} & \mathbf{K}_*^T\\
    \mathbf{K}_* & \mathbf{K}_{**}
    \end{bmatrix}
    \right)\,,
\end{equation}
where $\mathbf{K}$, $\mathbf{K}_{**}$, $\mathbf{K}_{*}$ are the train, test and test/train Gram matrices, respectively.
The posterior distribution of $\mathbf{f}_*$, obtained by conditioning the joint distribution on the observed data, is also Gaussian: $\mathbf{f}_* | \mathbf{G}, \mathbf{y}, \mathbf{G}^* \sim \mathcal{N}(\mathbf{\bar{m}},\mathbf{\bar{\Sigma}})$ with mean and covariance given by
\begin{align}
\mathbf{\bar{m}}&=\mathbf{K}_*(\mathbf{K}+\eta^2\mathbf{I})^{-1} \mathbf{y}\,, \\
\mathbf{\bar{\Sigma}}&=\mathbf{K}_{**}-\mathbf{K}_* (\mathbf{K}+\eta^2 \mathbf{I})^{-1}\mathbf{K}_*^T\,.
\end{align}
The mean of the posterior distribution is used as a predictor, and predictive uncertainties can be obtained through the covariance matrix. Figure \ref{fig:gp} gives a visual representation of the GP regression for graph inputs.

The predictive performance of GP regression strongly depends on the underlying kernel function $k$. Most importantly, this kernel function has to be positive definite. In the next section, we introduce our positive definite graph kernel based on optimal transport.

\section{SWWL GRAPH KERNEL} \label{section:swwl_for_gp}
The proposed methodology relies on the sliced Wasserstein distance \citep{sliced_barycenter} and the continuous Weisfeiler-Lehman (WL) iterations \citep{wwl} in order to build a positive definite kernel over graphs. 
The different steps are detailed after the definition of the kernel.

\begin{definition}[SWWL kernel]
\label{def:swwl}
Let $P\geq 1$ be the number of projections, $Q\geq 2$ be the number of quantiles, and $H\geq 0$ be the number of WL iterations. The SWWL kernel (illustrated by Figure \ref{fig:swwl}) is defined for any $G, G'\in \mathcal{G}$ by
\begin{equation} \label{eq:swwl_def}
k_{\mathrm{SWWL}}(G,G') = \exp \left(-\gamma \widehat{SW}^2_{2,P,Q}(\mu_G, \mu_{G'}) \right)\,,
\end{equation}
where 
$\mu_G = |V|^{-1} \sum_{u \in V } \delta_{\mathbf{E}^G_u}$
is the empirical distribution associated with the continuous WL embedding
$\mathbf{E}^G=(\mathbf{E}^G_u)_{u \in V}$ of $G$ with $H$ iterations (see Definition~\ref{def:wle})
and $\gamma> 0$ is a precision parameter.
\end{definition}

Our SWWL kernel has several key properties.
\begin{property}
\label{property1}
  There exists an explicit feature map $\bs{\phi}$ in a $PQ$-dimensional space (see Eq.\,\eqref{eq:PQE}) such that the SWWL kernel can be rewritten as
\begin{equation} \label{eq:swwl_phi}
k_{\mathrm{SWWL}}(G,G') = \exp \left( -\gamma ||\bs{\phi}(\mu_G)-\bs{\phi}(\mu_{G'})||^2_{2}\right)\,.  
\end{equation}
\end{property}

Eq.\,\eqref{eq:swwl_phi} shows that building the Gram matrix can be broken down into two parts: (1) computing the embeddings $\bs{\phi}(\mu_G)$ and (2) assembling their pseudo-distance matrix. Importantly, the time complexity of the latter step is independent of the number of graph nodes.

\begin{property} \label{prop:complexity_swwl}
Let $\delta$ be the average degree of nodes
and $n$ be the average number of nodes. The time complexity for assembling the Gram matrix of the kernel given by Eq.\,\eqref{eq:swwl_def} is
\begin{align*}
    \mathcal{O}(NH\delta n + NPn(\log(n)+H) +N^2PQ)\,.
\end{align*}
The space complexity is $\mathcal{O}(NPQ)$.
\end{property}

\begin{property} \label{prop:swwl_pd}
The SWWL kernel is positive definite.
\end{property}
The proofs are given in the Supplementary Material (the positive definiteness result relies on the work of \cite{meunier_sliced_pd}).
We now detail the construction of the feature map $\bs{\phi}$ involved in Eq.\,\eqref{eq:swwl_phi}.

\paragraph{Continuous Weisfeiler-Lehman embeddings.}
The WL embeddings were originally proposed for graphs having nodes with discrete/categorical labels, but they were extended to continuous attributes by \cite{wwl}.
\begin{definition}[Continuous WL embeddings]
\label{def:wle}
Let $G = (V, E, w, \mathbf{F})$ be a graph with attributes $\mathbf{F}=(\mathbf{F}_u)_{u \in V}$, $ \mathbf{F}_u \in \mathbb{R}^d$. The continuous Weisfeiler-Lehman iterations are defined recursively for $h\in \mathbb{N}$ by
\begin{equation} \label{eq:cWLit}
\mathbf{F}^{(h+1)}_u = \frac{1}{2} \bigg( \mathbf{F}^{(h)}_u + \frac{1}{\mathrm{deg}(u)}\sum_{v\in \mathcal{N}(u)} w({u,v}) \mathbf{F}^{(h)}_v \bigg)\, ,
\end{equation}
with  $\mathbf{F}_u^{(0)}= \mathbf{F}_u$ for $u \in V$.
Given a number of iterations $H\geq 0$, the continuous WL node embedding of the graph $G$ is the concatenation of the WL iterations at steps $0, 1, \ldots, H$ denoted by
$\mathbf{E}^G =( \mathbf{E}^G_u )_{u \in V} $, $\mathbf{E}^G_u \in \mathbb{R}^{(H+1)d}$.
\end{definition}
\paragraph{Optimal transport distances.}
Once the continuous WL embeddings $\mathbf{E}^G$ have been obtained for all graphs, their empirical distributions can be considered and the Wassertein distances can be computed between all pairs of empirical distributions to assemble a kernel. We first recall the definition of the Wasserstein distance for arbitrary measures on $\mathbb{R}^s, s\geq 1$, but we will see they do not lead to positive definite kernels.

\begin{definition}[Wasserstein distance]
\label{def:wasserstein}
Let $s \geq 1$ be an integer, $r \geq 1$ be a real number, and $\mu, \nu$ be two probability measures on $\mathbb{R}^s$ having finite moments of order $r$. The $r$-Wasserstein distance is defined as
\begin{equation}
W_r(\mu, \nu) := \bigg( \inf\limits_{\pi \in \Pi(\mu, \nu)} \int\limits_{\mathbb{R}^s\times\mathbb{R}^s} ||\mathbf{x}-\mathbf{y}||^r d\pi(\mathbf{x},\mathbf{y}) \bigg)^{\frac{1}{r}}\,,  
\end{equation}
where $\Pi(\mu, \nu)$ is the set of all probability measures on $\mathbb{R}^s\times\mathbb{R}^s$ whose marginals w.r.t.
the first and second variables are respectively $\mu$ and $\nu$ and $||\cdot||$ stands for the Euclidean norm on $\mathbb{R}^s$.
\end{definition}

The Wasserstein distance between two empirical measures can be computed with linear programming in $\mathcal{O}(n^3 \log(n))$ or accelerated with entropy regularization (e.g. Sinkhorn iterations \citep{sinkhorn}) in $\mathcal{O}(n^2 \log(n))$. Distance substitution kernels \citep{distance_substitution} offer a natural way to build kernel functions between empirical distributions.
Unfortunately, it is not possible to design positive definite kernels by plugging the Wasserstein distance into such kernels when the dimension $s$ of the space $\mathbb{R}^s$ is greater than one \citep{computational_ot}.
%
%
To circumvent this issue, we rely on the sliced Wasserstein distance \citep{sliced_barycenter}, which averages one-dimensional Wasserstein distances between the distributions projected on the unit sphere. First, the complexity is reduced because the $1d$-Wasserstein distance can be computed in $\mathcal{O}(n \log(n))$, and second, the corresponding substitution kernel is positive definite.


\begin{definition}[Sliced Wasserstein distance]
\label{def:sw}
Let $s\geq 1$, $r\geq 1$. The $r$-sliced Wasserstein distance is defined as
\begin{equation} \label{eq:sliced_wasserstein}
SW_r(\mu,\nu) := \bigg(\int\limits_{\mathbb{S}^{s-1}} W_r(\theta^*_\sharp \mu, \theta^*_\sharp \nu)^r d\sigma(\theta) \bigg)^{\frac{1}{r}}\,,
\end{equation}
where $\mathbb{S}^{s-1}$ is the $(s-1)$-dimensional unit sphere, $\sigma$ is the uniform distribution on $\mathbb{S}^{s-1}$,  $\theta^* : \mathbf{x} \mapsto \langle \mathbf{x}, \bs{\theta} \rangle$ the projection function of $\mathbf{x} \in \mathbb{R}^s$ in the direction $\bs{\theta} \in \mathbb{S}^{s-1}$ and $\theta^*_\sharp \mu$ the push-forward measure of $\mu$ by $\theta^*$.
\end{definition}

In practice, Monte Carlo estimation is performed by sampling $P$ directions $\bs{\theta}_1, \ldots, \bs{\theta}_P$ uniformly on $\mathbb{S}^{s-1}$, with controlled error bounds (\cite{mc_estimate, mc_estimate2}). 
Although questioned in recent approaches (\cite{max_sliced}), Monte Carlo estimation is a common approach that leads to good results in practice. Alternatively, quasi Monte Carlo methods could be used (in particular for low dimensional problems).

\paragraph{Projected quantile embeddings.}
The computation of the $1d$-Wasserstein distance between empirical measures usually consists in sorting the projected points and then summing a power of the Euclidean distances between values at the same ranks. Here, we propose instead to use $Q \ll n$ equally-spaced quantiles, that do not depend on the distribution sample size $n$ (in practice, data points are sorted and if the quantile lies between two data points, a linear interpolation is used). This strategy differs from usual implementations \citep{pot}, where the grid of quantiles is chosen according to each pair of input distributions. More precisely, let $0 = t_1 < \dots < t_\ell < \dots < t_Q = 1$ be a sequence of $Q$ equally-spaced points in $[0,1]$, and let $\bs{\theta} \in \mathbb{S}^{s-1}$ be a projection direction. The $Q$-quantiles estimation of the $1d$-Wasserstein distance between the 
push-forward measures $\mu_\theta := \theta^*_\sharp \mu$ and $\nu_\theta := \theta^*_\sharp \nu$ writes

\begin{equation}\label{eq:estimation_1d_wasserstein}
    \tilde{W}_{r,Q}(\mu_\theta, \nu_\theta) = \bigg( \frac{1}{Q} \sum\limits_{\ell=1}^{Q} \left|F^{-1}_{\mu_\theta}\left(t_\ell\right) - F^{-1}_{\nu_\theta}\left(t_\ell\right)\right|^r   \bigg)^{\frac{1}{r}}
\end{equation}
where $F_\mu(x)=\mu((-\infty, x]), x\in \mathbb{R}$ is the cumulative distribution function and $F_\mu^{-1}(t) = \inf \{x\in \mathbb{R} : F_\mu(x)\geq t\}, t\in [0,1]$ is the inverse cumulative distribution function. 
The sliced Wasserstein distance given by Eq.\,\eqref{eq:sliced_wasserstein} is finally estimated as follows:
\begin{equation} \label{eq:MC_wasserstein}
\widehat{SW}_{r,P,Q}(\mu, \nu) = \bigg( \frac{1}{P} \sum_{p=1}^{P} \tilde{W}_{r,Q}(\mu_{\theta_p}, \nu_{\theta_p})^r \bigg)^{\frac{1}{r}}\,,
\end{equation}
where $\bs{\theta}_1, \ldots, \bs{\theta}_P$ denote $P$ projection directions uniformly drawn on $\mathbb{S}^{s-1}$. By combining Eqs.\,\eqref{eq:estimation_1d_wasserstein} and \eqref{eq:MC_wasserstein}, this estimate can be rewritten as
\begin{equation} \label{eq:PQE_wasserstein}
\widehat{SW}_{r,P,Q}(\mu, \nu) = ||\bs{\phi}(\mu)-\bs{\phi}(\nu)||_{r}\,,
\end{equation}
where $||.||_{r}$ is the $r$-norm in $\mathbb{R}^{PQ}$, and $\bs{\phi}$ is the explicit feature map
\begin{equation} \label{eq:PQE}
    \bs{\phi}_{p + P(q-1)}(\mu) = (PQ)^{-1/r} F^{-1}_{\mu_{\theta_p}}(t_q)
\end{equation}
for $p = 1, \dots, P$, $q = 1, \dots, Q$. This feature map is called the \emph{projected quantile embedding}. It provides a $PQ$-dimensional representation of any probability distribution $\mu$ on $\mathbb{R}^s$. In Definition~\ref{def:swwl} the probability distributions of interest are the empirical distributions associated with the continuous WL embeddings
$\mathbf{E}^G=(\mathbf{E}^G_u)_{u \in V}$ of graphs $G \in {\mathcal G}$ with $H$ iterations, so that $s=H(d+1)$. We, therefore, get Property~\ref{property1}.

\section{EXPERIMENTS} \label{section:exp}
In our experiments \footnote{Code: \url{https://gitlab.com/drti/swwl/}}, we consider two different tasks:
\begin{itemize}[topsep=0pt]
\itemsep0em 
    \item Classification of small graphs with less than 100 nodes corresponding to molecules. This serves as a necessary validation of our kernel with respect to state-of-the-art kernels, and as an illustration of its reduced complexity. 
     \item Gaussian process regression for large graphs from mesh-based simulations in computational fluid dynamics and mechanics. Here we demonstrate that the SWWL kernel can easily handle more than $10^5$ nodes in a few seconds or minutes.
\end{itemize}
Details about the datasets and in particular their graph characteristics are given in Table \ref{tab:data}.
\begin{table*}[h]
\caption{Summary of the datasets. $(*)$: fixed number of nodes and adjacency structure, CM: coarsened meshes.}
\label{tab:data}
\centering
\begin{tabular}{ccccccc} 
 \toprule
  Dataset & Num graphs & Mean nodes & Mean edges & Attributes & Scalars & Task (classes) \\
 \midrule
 BZR & 405 & 35.7 & 38.4 & 3 & 0 & Classif(2) \\ 
 COX2 & 467 & 41.2 & 43.5 & 3 & 0 & Classif(2) \\
 PROTEINS & 1113 & 39.1 & 72.8 & 1 & 0 & Classif(2)\\
 ENZYMES & 600 & 32.6 & 62.1 & 18 & 0 & Classif(6)\\
 Cuneiform & 267 & 21.27 & 44.8 & 5 & 0 & Classif(30)\\
 \midrule \midrule
 Rotor37$ ^*$ & 1000+200 & 29773 & 77984 & 3 & 2 & Regression\\
 Rotor37-CM & 1000+200 & 1053.8 & 3097.4 & 3 & 2 & Regression\\
 Tensile2d & 500+200 & 9425.6 & 27813.8 & 2 & 6 & Regression \\
 Tensile2d-CM & 500+200 & 1177.4 & 3159.8 & 2 & 6 & Regression \\
 AirfRANS & 800+200 & 179779.0 & 536826.6 & 2 & 2 & Regression \\
 AirfRANS-CM & 800+200 & 6852.8 & 19567.2 & 2 & 2 & Regression \\
 \bottomrule
\end{tabular}
\end{table*}

\begin{table*}[bp]
\setlength{\tabcolsep}{3pt}
\caption{Mean accuracy (\%) and standard deviation for 10 experiments of classification of small graphs.}
\label{tab:classif} 
\centering
\begin{tabular}{ccccccc} 
 \toprule
 & Kernel/Dataset & BZR & COX2 & PROTEINS & ENZYMES & Cuneiform \\
 \midrule
 OT-based & SWWL (ours) & \bf{85.43 $\pm$ 4.05} & 78.61 $\pm$ 5.87 & \bf{75.12 $\pm$ 5.99} & 66.67 $\pm$ 5.0 & 83.36 $\pm$ 4.32 \\ 
 & WWL & 84.43 $\pm$ 2.82 & 75.62 $\pm$ 6.43 & 74.85 $\pm$ 4.97 & \bf{70.33 $\pm$ 2.87} & 84.62 $\pm$ 6.78 \\
 & FGW & \bf{85.41 $\pm$ 3.14} & 76.05 $\pm$ 7.98 & 71.79 $\pm$ 3.61 & 67.83 $\pm$ 2.36 & 80.85 $\pm$ 8.06\\
 & RPW & \bf{85.42 $\pm$ 2.41} & 77.98 $\pm$ 5.54 & 71.42 $\pm$ 5.10 & 52.0 $\pm$ 6.94 & \bf{91.00  $\pm$ 8.36}\\
 Non-OT-based & PK & 80.96 $\pm$ 4.79 & 78.21 $\pm$ 7.41 & 69.54 $\pm$ 4.90 & 68.5 $\pm$ 5.13 & - \\
 & GH & 82.44 $\pm$ 4.98 & \bf{79.49 $\pm$ 6.04} & 71.97 $\pm$ 2.44 & 43.5 $\pm$ 3.91 & - \\
 \bottomrule
\end{tabular}
\end{table*}

\begin{table*}[h]
\caption{Computation times (s) needed to build the distance/Gram matrices of small graphs for all hyperparameters. When embeddings are pre-computed, we give both embedding times + distance times.}
\label{tab:classif_times} 
\centering
\begin{tabular}{ccccccc} 
 \toprule
 & Kernel/Dataset & BZR & COX2 & PROTEINS & ENZYMES & Cuneiform \\
 \midrule
 OT-based & SWWL (ours) & \bf{0.7 + 0.1} & \bf{0.6 + 0.1} & \bf{1.5 + 0.6} & \bf{1.1 + 0.2} & \bf{1.3 + 0.1} \\ 
 & WWL & 0.3 + 97 & 0.3 + 131.2 & 0.7 + 803 & 0.5 + 220 & 0.9 + 34\\
 & FGW & 0.6 + 714 & 0.7 + 842 & 1.6 + 7882 & 0.9 + 1381 & 0.4 + 145 \\
 & RPW & 35 + 5 & 40 + 7 & 240 + 40 & 220 + 40 & -\\
 Non-OT-based & PK & 10 & 13 & 52 & 53 & -\\
 & GH & 77 & 108 & 3998 & 230 & - \\
 \bottomrule
\end{tabular}
\end{table*}

\subsection{Classification}

\paragraph{Experimental setup.}
The proposed SWWL kernel is compared with 5 existing graph kernels, namely, WWL \citep{wwl}, Fused Gromov Wasserstein (FGW) \citep{fgw}, Restricted Projected Wasserstein (RPW) \citep{a_simple_way_to_learn_metrics_btw_attributed_graphs}, Propagation (PK) \citep{propag} and Graph Hopper (GH) \citep{hopper}. 
Five datasets are taken from the TUD benchmark \citep{TUDatasets}: BZR, COX2, PROTEINS, ENZYMES, Cuneiform. We only keep continuous attributes for the nodes and ignore categorical information except for Cuneiform where we use the same setup as proposed by \cite{a_simple_way_to_learn_metrics_btw_attributed_graphs} with fused attributes. 

Following the same approach as in previous work, we use a SVM classifier with the (indefinite) kernel $k(G,G') := \exp ( -\gamma \mathcal{D}(G,G'))$  where $\mathcal{D}$ is the (pseudo)distance defined by the method at hand. There are $3$ hyperparameters to set: the number of WL iterations $H \in \{0,1,2,3\}$, the kernel precision $\gamma \in \{10^{-4} , 10^{-3} , \ldots , 10^{-1}\}$, and the SVM regularization
parameter $C\in \{10^{-3}, 10^{-2}, \ldots, 10^3\}$. In the case of the SWWL kernel, we arbitrarily set the number of projections to $P = 20$, and the number of quantiles to $Q = 20$. It has been found that their influence is negligible on these small datasets. A more detailed study in the case of regression tasks with large graphs is carried out in Section \ref{section:regression}.

To train and evaluate all baselines, we rely on nested cross-validation as in \citet{wwl}. We use stratified $10$-fold train/test split in the outer loop. For each split, the model hyperparameters are tuned by a cross-validation scheme with an inner stratified $5$-fold defined on the current train set. We then compute the accuracy for the current test set and finally average the results obtained for all the $10$ outer splits. The same random splits and seeds are used for all the considered methods in order to ensure reproductibility and fair comparisons. Results are reported in Table \ref{tab:classif}.

\paragraph{Discussion about the results.}
As Table \ref{tab:classif} shows, our SWWL kernel achieves similar results for graph classification as other state-of-the-art kernels. It is worth emphasizing again that the SWWL kernel is not intended to outperform other methods here, but rather to be a positive-definite variant of the WWL kernel with drastically reduced complexity. Remark that due to the small sample size and high class imbalance in these datasets, the results strongly depend on the initial train/test split, thus explaining the observed high variance.


Time-based comparisons of methods are given in Table \ref{tab:classif_times}. They should be taken with hindsight, especially when considering that they do not all rely on the same backend. Computation times for the RPW kernel are taken from the original paper, since our own implementation was much slower. In summary, it can be seen that the SWWL kernel largely outperforms other methods in terms of computation times thanks to its explicit unsupervised embeddings.

\subsection{Regression} \label{section:regression}
We now apply the SWWL kernel to three challenging high-dimensional regression problems from computational fluid and solid mechanics \footnote{Datasets: \url{https://plaid-lib.readthedocs.io/en/latest/source/data_challenges.html}. See also the Supplementary Material.}.

\smallskip

\textit{Rotor37 dataset}. The NASA rotor 37 case \citep{Rotor37Dataset} serves as a prominent example of a transonic axial-flow compressor rotor widely employed in computational fluid dynamics research. 
This dataset is made of 3D compressible steady-state Reynold-Averaged Navier-Stokes (RANS) simulations that model external flows \citep{ameri2009nasa}. The inputs of the simulations are given by the finite element mesh of a 3D compressor blade (graphs with $\sim 30000$ nodes), and two scalar physical parameters corresponding to the rotational speed and the input pressure. The scalar output of the problem is the isentropic efficiency, which is computed by solving the boundary value problem with the finite volume method  using the \texttt{elsA} software \citep{cambier2011overview}. 


\smallskip

\textit{Tensile2d dataset}. We consider a two-dimensional problem in solid mechanics introduced by \cite{Tensile2dDataset}. The input geometries consist of 2D squares with two half circles that have been cut off in a symmetrical manner. Pieces are subject to a uniform pressure field over the upper boundary and the material of the structures is modeled by a nonlinear elasto-viscoplastic law. The inputs of the problem are given by the mesh of the geometry (graphs with $\sim 10000$ nodes), and six scalar parameters that correspond to the material parameters and the input pressure applied to the upper boundary. The scalar output of the problem is the maximum Von Mises stress across the domain, computed numerically with the finite element method using the \texttt{Zset} software \cite{garaud2019z}.

\smallskip

\textit{AirfRANS dataset}. This dataset consists of RANS simulations over 2D NACA airfoils with geometric variabilities in a subsonic flight regime. The inputs of the problem are given by the finite element mesh of the domain surrounding the airfoil (graphs with more than $10^5$ nodes), and two scalar inputs, the angle of attack and the inlet velocity. The scalar output of the problem is the drag coefficient. More details about the distributions of the training and test sets can be found in the work of \cite{naca}.

\begin{figure}[h]
\begin{center}
\includegraphics[height=2.2cm, keepaspectratio]{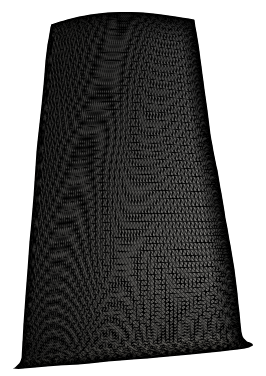}
\includegraphics[height=2.2cm, keepaspectratio]{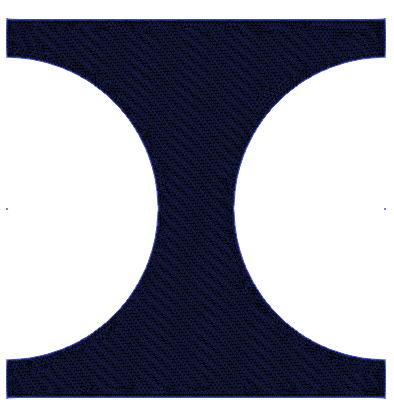}
\includegraphics[height=2.2cm, keepaspectratio]{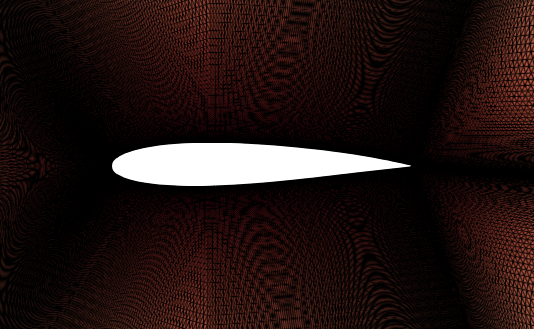}
\includegraphics[height=2.2cm, keepaspectratio]{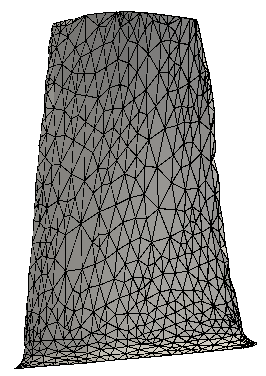}
\includegraphics[height=2.2cm, keepaspectratio]{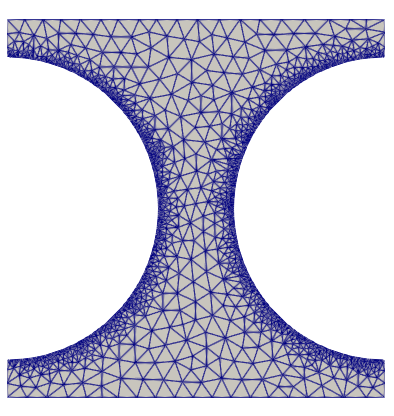}
\includegraphics[height=2.2cm, keepaspectratio]{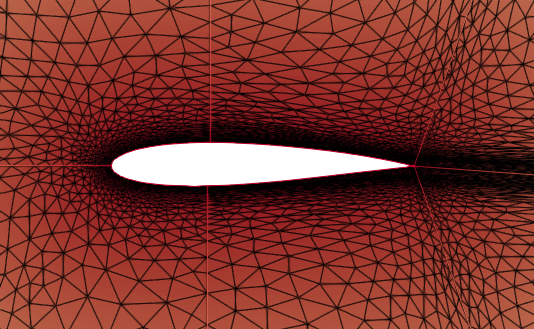}
\end{center}
\caption{Top row: meshes from the \texttt{Rotor37}, \texttt{Tensile2d} and \texttt{AirfRANS} datasets (from left to right). Bottom row: coarsened versions of the meshes.}
\label{fig:meshes}
\end{figure}

As summarized in Table \ref{tab:data}, the input meshes are made of a large number of nodes. While our SWWL kernel is able to deal with such large graphs, we have observed numerically that no alternative graph kernel is tractable in this setting. For this reason, we also consider a degraded version of each dataset with a largely reduced number of nodes. Such smaller dimensional problems are obtained with coarsened meshes (CM) using the MMG remesher \citep{mmg}, while the scalar inputs and outputs are unchanged. The associated datasets are referred to \texttt{Rotor37-CM}, \texttt{Tensile2d-CM}, and \texttt{AirfRANS-CM}. Figure \ref{fig:meshes} shows samples from the datasets and their coarsened variants.

\paragraph{Choice of the number of iterations.}
Usually it is recommended to choose a small number of iterations to avoid an oversmoothing phenomenon (\cite{xu2018rpz}), but the impact of this hyperparameter is largely overlooked in the literature. Since we work here with large graphs, we propose to investigate the incorporation of skips during the WL iterations in order to visit larger neighborhoods. The step $T$ is chosen to be approximately $\sqrt{n}$ and we keep only iterations $0$, $T$, $2T$ and $3T$: the resulting kernel is called SWWL($\sqrt{n}$) in the following, in contrast to SWWL($1$) for the original one with iterations $0,\ldots,3$. However note that computing iterations with $T=O(\sqrt{n})$ changes the complexity of SWWL embeddings from $\mathcal{O}(NH\delta n + NPn\log(n))$ to $\mathcal{O}(N\delta n \sqrt{n} + NPn\log(n))$. The default number of projections for both kernels is set to 50 and the default number of quantiles to 500.

\paragraph{Experimental setup.}
For all the considered problems, the train and test sets have been generated by a space-filling design of experiments, including for the inputs given as graphs, since the mesh geometry is actually parameterized beforehand. The training of the GP is repeated $5$ times, and we report the mean and standard deviations of the evaluation metrics. The kernel function of the GP is chosen as a tensorized kernel of the form
\begin{equation} \label{eq:tensorized}
     k(\mathfrak{X}, \mathfrak{X}') := \sigma^2 k_{\mathrm{SWWL}}(G,G') \prod_{\ell=1}^{m} c_{5/2}(|s_\ell-s'_\ell|)
\end{equation}
 where $\mathfrak{X} = (G, s_1,\ldots, s_m)$ and $\mathfrak{X}' = (G', s'_1,\ldots, s'_m)$ denote the input graphs and the $m$ input scalars $s_1,\ldots,s_m$, $c_{5/2}$ is the Matérn-$5/2$ covariance function, and $\sigma^2$ is a variance parameter. The lengthscale parameters of the SWWL and Matérn-5/2 kernels are optimized simultaneously by maximizing the marginal posterior log-likelihood following the work of \cite{rgasp} (see the Supplementary Material).
\begin{table*}[h]
\setlength{\tabcolsep}{3pt}
\caption{Mean RMSE and standard deviation for 5 experiments of Gaussian process regression. Missing values correspond to intractable experiments due to memory and/or CPU usage.}
\label{tab:regression}
\centering
\begin{tabular}{ccccccc} 
 \toprule
 Kernel/Dataset & \texttt{Rotor37} & \texttt{Rotor37-CM} & \texttt{Tensile2d} & \texttt{Tensile2d-CM} & \texttt{AirfRANS} & \texttt{AirfRANS-CM} \\
 & x10\textsuperscript{-3} & x10\textsuperscript{-3} & x1 & x1 & 
 x10\textsuperscript{-4}&
 x10\textsuperscript{-4}\\
 \midrule
 SWWL(1) & 1.44 $\pm$ 0.07 & \bf{3.49 $\pm$ 0.15} & \bf{0.89 $\pm$ 0.01} & \bf{1.51 $\pm$ 0.01} & 7.56 $\pm$ 0.36 & 9.63 $\pm$ 0.54 \\
  SWWL($\sqrt{n}$) & \bf{1.36 $\pm$ 0.05} & 3.64 $\pm$ 0.10 & \bf{0.90 $\pm$ 0.01} & 1.57 $\pm$ 0.03 & \bf{7.29 $\pm$ 0.41} & \bf{8.07 $\pm$ 0.34}\\
  WWL & - & \bf{3.51 $\pm$ 0.00} & - & 6.46 $\pm$ 0.00 & - & 14.4 $\pm$ 0.80 \\
  PK & - & 4.18 $\pm$ 0.39 & - & 6.03 $\pm$ 4.58 & - & 8.94 $\pm$ 2.31 \\
 \bottomrule
\end{tabular}
\end{table*}


\begin{table*}[h]
\setlength{\tabcolsep}{3pt}
\caption{Embedding and distance computation times needed for a single Gram matrix. When embeddings are pre-computed, we give both embedding times + distance times. (*): Parallelized over $100$ processes.}
\label{tab:regression_times}
\centering
\begin{tabular}{ccccccc} 
 \toprule
 Kernel/Dataset & \texttt{Rotor37} & \texttt{Rotor37-CM} & \texttt{Tensile2d} & \texttt{Tensile2d-CM} & \texttt{AirfRANS} & \texttt{AirfRANS-CM} \\
 \midrule
  SWWL(1) & \bf{1min + 11s} & \bf{4s + 11s} & \bf{11s + 4s} & \bf{2s + 4s}& \bf{5min + 7s} & \bf{15s + 7s} \\ 
  SWWL($\sqrt{n}$) & 54min + 11s & 20s + 11s& 5min + 4s& 10s + 4s& 3h52min + 7s& 7min + 7s\\ 
  WWL & - & 13min (*)  & -  & 6min (*) & - & 8h (*) \\
  PK & - & 1min & - & 2min & - & 15min\\ 
 \bottomrule
\end{tabular}
\end{table*}

\paragraph{Discussion about the results.} 
We first observe in Table \ref{tab:regression} that adding the information of skipped iterations decreases the RMSE on \texttt{AirfRANS-CM} and \texttt{AirfRANS}, but it does not seem to significantly improve performance on other datasets. Such a decrease in RMSE however does not seem to justify the important rise in computation time for the embeddings. Second, we achieve lower RMSE for SWWL than for WWL and PK on all CM datasets. Table \ref{tab:regression_times} also illustrates that SWWL outperforms other methods in terms of computation times even when the number of WL iterations is large. But on these coarsened datasets, even though they make other kernels tractable, we achieve much lower accuracies when compared to the original ones, meaning that the naive coarsening strategy is not efficient. Taking the complete data and doing an efficient dimension reduction with SWWL is undoubtedly a better strategy.

\begin{figure}[ht]
\begin{center}
\includegraphics[scale=0.42]{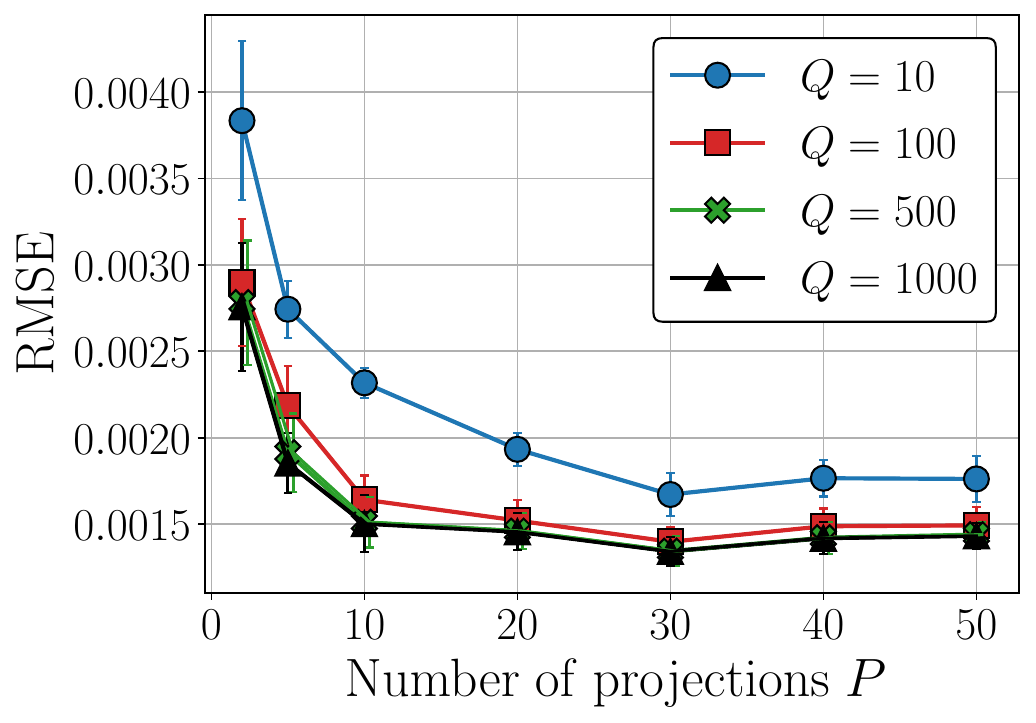}\\
\end{center}
\caption{Impact of number of projections and quantiles on RMSE for GP regression for \texttt{Rotor37}.}
\label{fig:rotor37_q2_projections}
\end{figure}

Finally, we also investigate the impact of the number of projections and the number of quantiles on the SWWL kernel. Figure \ref{fig:rotor37_q2_projections} shows that on the \texttt{Rotor37} dataset, the mean RMSE decreases and the standard deviation diminishes when the number of projections and quantiles grow, as expected. However, above 20 projections and $100$ quantiles the RMSE score decreases very little, and we observe the same behavior for all other datasets. This means that in general an embedding of size $20 \times 100$ is sufficient to achieve good performance.

\paragraph{Implementation and computing infrastructure.}
We leverage an available Python implementation of continuous WL iterations \citep{pytorch_geom}, and GP regression in R \citep{rgasp}. The FGW distances are computed with the POT library \citep{pot}. The PK and GH kernels are implemented in the Grakel library \citep{grakel}, and we use the original implementation of RPW \citep{a_simple_way_to_learn_metrics_btw_attributed_graphs}. All our analyses were performed on a shared server running Linux CentOS 7.9.2009, with 1 CPU (Intel Xeon Gold 6342 @ 2.80GHz) using 4 cores, and a total of 21,6 GB of RAM.

\section{CONCLUSION}
The Sliced Wasserstein Weisfeiler-Lehman is a new positive definite graph kernel with drastically reduced complexity compared to existing graph kernels. It is more efficient for graphs where attributes are of paramount importance compared to the adjacency structure. We demonstrate its efficiency, with equivalent performance to state-of-the-art kernels for common benchmarks of small molecules. We also propose for the first time Gaussian process regression of large-scale meshes used in computational physics where the size of graphs usually limits the use of existing kernels. Our kernel can be viewed as an explicit unsupervised embedding, with complexity related to the number of projections and quantiles used in the sliced Wasserstein distance. Our experiments show that few of them are actually sufficient to encapsulate information, thus achieving an explicit and efficient dimension reduction.

We only considered here supervised learning tasks, but our kernel can be plugged into any other kernel-based method, and may thus be of particular interest for graph clustering with kernel k-means, or space-filling designs with the maximum mean discrepancy.

A remaining question is the impact of the unsupervised WL iterations. A more supervised version with an optimization phase may help increase the representation capacity, at the expense of non-negligible additional training cost, and such trade-off should be further studied. Finally, our kernel can only handle scalar outputs, and future work may include an extension to vector fields, since in computational physics the outputs obtained by simulation are often of this form.

\subsubsection*{Acknowledgements}
This work was partially supported by the Agence Nationale de la Recherche through the SAMOURAI (Simulation Analytics and Metamodel-based solutions for Optimization, Uncertainty and Reliability AnalysIs) project under grant ANR20-CE46-0013 and the EXAMA (Methods and Algorithms at Exascale) project under grant ANR-22-EXNU-0002.
\bibliography{b}


\appendix
\onecolumn
\aistatstitle{Supplementary Material for Gaussian process regression with Sliced Wasserstein Weisfeiler-Lehman graph kernels}
\section{Proofs}
\subsection{Positive definiteness of SWWL}
In this section we give a proof of Property \ref{prop:swwl_pd} (The SWWL kernel is positive definite).
We first recall useful properties about Hilbertian pseudo-distances and their link with positive definite kernels. 

\begin{definition}[Hilbertian (pseudo)-distance \citep{hilbertian}]
\label{def:hilbertian_d}
    A pseudo-distance $d$ defined on a set $X$ is said to be \emph{Hilbertian} if there exists a Hilbert space $\mathcal{H}$ and a feature map $\phi: X \rightarrow \mathcal{H}$ such that for any pair $x,x'$ in $X$, $d(x,x') = ||\phi(x)-\phi(x')||_{\mathcal{H}}$.
\end{definition}

All usual kernels substituted with distances are positive definite if and only if the distance is Hilbertian as the following property shows.

\begin{property} [\cite{meunier_sliced_pd}]
\label{prop:hilbertian_equiv}
Introducing $\langle x,x'\rangle^{x_0}_d = \langle \phi(x)-\phi(x_0), \phi(x')-\phi(x_0) \rangle_\mathcal{H}$, the following statements are equivalent:
\begin{itemize}[topsep=0pt]
\itemsep0em 
\item $d$ is a Hilbertian pseudo-distance.
\item $k_{lin}(x,x')= \langle x,x'\rangle^{x_0}_d $, $(x,x')\in X^2$ is positive definite for all $x_0\in X$.
\item $k_{poly} (x,x') = \left( c+\langle x, x'\rangle^{x_0}_d \right)^{l}$, $(x,x')\in X^2$ is positive definite for all $x_0 \in X$, $c\geq 0$, $l\in \mathbb{N}$.
\item $k(x,x') = \exp( -\gamma d^{2\beta}(x,x'))$, $(x,x')\in X^2$ is positive definite for all $\gamma \geq 0$, $\beta\in [0,1]$.
\end{itemize}
\end{property}

It is possible to show that Wasserstein distances are not Hilbertian, but that sliced Wasserstein distances are Hilbertian.

\begin{property} [\cite{computational_ot}]
\label{prop:wasserstein_not_hilbertian}
$\sqrt{W_1}$ and $W_2$ are not Hilbertian distances in $\mathbb{R}^s$ when $s\geq2$.
\end{property}

\begin{property} [\cite{meunier_sliced_pd}]
\label{prop:sw_hilbertian}
$\sqrt{SW_1}$ and $SW_2$ are Hilbertian distances in $\mathbb{R}^s$ for all $s\geq2$. 
\end{property}


We can now prove the positive definiteness of SWWL.
\begin{proof}
We recall the form of the estimated sliced Wasserstein distance between two distributions $\mu$ and $\nu$ with $P$ projections and $Q$ quantiles:
$\widehat{SW}_{r,P,Q}(\mu, \nu) = ||\phi(\mu)-\phi(\nu)||_{r}$
where $||.||_{r}$ is the $r$-norm in $\mathbb{R}^{PQ}$ and  
$\bs{\phi}_{p + P(q-1)}(\mu) = (PQ)^{-1/r} F^{-1}_{\mu_{\theta_p}}(t_q)$ for $p = 1, \dots, P$ and $q = 1, \dots, Q$.
$\widehat{SW}_{r,P,Q}$ is a Hilbertian pseudo-distance \ref{def:hilbertian_d} as we built an explicit feature map $\phi$ to $\mathcal{H} = \mathbb{R}^{PQ}$.\\
If we deal with empirical measures with maximal support size $\hat{n}$ (supposed to be finite), $\widehat{SW}_{r,P,Q}$ is not necessarily a distance when $Q<\hat{n}$ as points from some measures can be lost by retaining only $Q$ quantiles.  

Using Property \ref{prop:hilbertian_equiv}, we obtain that:
$\mu, \nu \mapsto \exp \left(-\gamma \widehat{SW}^2_{2,P,Q}(\mu, \nu) \right)$ is a positive definite kernel for $\gamma>0$.

Finally, this property still holds if we restrict the space of measures to be 
$\{ \mu_G = |V|^{-1} \sum\limits_{u \in V} \delta_{\mathbf{E}^G_u}, G \in \mathcal{G}\}$, the empirical distributions associated with the continuous WL embeddings of graphs in $\mathcal{G}$.

We conclude that for $\gamma>0$, $k_{\mathrm{SWWL}}(G,G') = \exp \left(-\gamma \widehat{SW}^2_{2,P,Q}(\mu_G, \mu_{G'}) \right)$ is a positive definite kernel.
\end{proof}

If we take $Q\geq max\{|V| , G\in \mathcal{G}\}$ and $P\rightarrow +\infty$, $\widehat{SW}_{r,P,Q}$ is then exactly the sliced Wasserstein distance for empirical measures. 

Remark that our approximation of sliced Wasserstein distances is different from the one proposed in \cite{meunier_sliced_pd}. We actually used a fixed number of quantiles $Q$ and we look at equally spaced values, whereas for them positions are drawn uniformly at random, which does not guarantee a complete recovery of the original distribution even if the number of draws is equal to its size.
\subsection{Complexity of SWWL}
In this section we give a proof of Property \ref{prop:complexity_swwl} (Complexity of SWWL).

\begin{proof}
$N$ denotes the number of graphs, $n$ the average number of nodes, $P$ the number of projections and $Q$ the number of quantiles.

The first step is the computation of WL iterations. For a graph of size $n$, one WL iteration propagates information of nodes to their neighbours in $O(n\delta)$ where $\delta$ is the mean number of neighbours.
Performing $H$ iterations for all the $N$ graphs is done in $O(NH\delta n)$.
The next step is to compute all $P$ projections for all graphs in $O(NPHn)$ and then to sort the 1-dimensional projected embeddings to obtain the $Q$ quantiles, which can be done for all graphs in $O(NPn\log(n))$.
Once projected quantile embeddings are computed for all graphs, the kernel is built by taking all pairwise distances between the embeddings in $O(N^2 PQ)$ (see Figure~\ref{fig:wwl}).\\
We conclude that the time complexity of SWWL is $O(NH\delta n + NPn(\log(n)+H)+N^2PQ)$.

\begin{figure}[h]
\begin{center}
\includegraphics[scale=0.25]{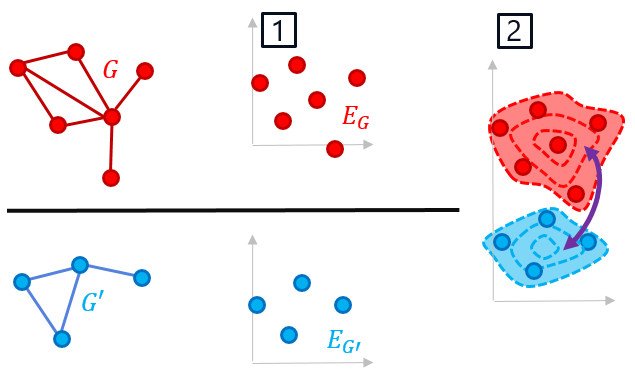}\\
\end{center}
\caption{Distance between embeddings of graphs seen as distributions. The black line represents operations that can be performed separately on each input $G_1, \ldots, G_N$. Step 1: graph embedding. Step 2: distance between distributions.}
\label{fig:wwl}
\end{figure}

For the spatial complexity, we need to store all the projected quantile embeddings in $O(NPQ)$. 
Remark that in theory, all embeddings do not need to be loaded simultaneously. During the computation of the distance matrix, we could only load  2 graphs in memory each time a distance is computed, resulting in a space complexity of $O(PQ)$. In practice, it is more convenient to keep all embeddings in memory in order to compute pairwise distances required for the Gram matrix at once.
\end{proof}

In comparison, using the Wasserstein distance would lead to a $O(N^2 n^3\log(n))$ time complexity and a $O(n^2)$ space complexity to build the Gram matrix, which can be prohibitive when the number of nodes is large.


\section{Optimal transport: definitions and properties}
Optimal transport, also known as Monge-Kantorovich transportation theory, is a mathematical framework that addresses the efficient transportation of mass or resources from one distribution to another. Optimal transport has found applications in various fields such as economics, physics, computer science, and machine learning. 
One of the theory's successes is to allow the definition of metrics/divergences between two probability measures with an interesting geometric interpretation. In this section we recall some basic definitions used in optimal transport. For a more exhaustive review, we refer the reader to \cite{computational_ot}. 

\begin{definition}[Push-forward operator](\cite{computational_ot})
Let $\mathcal{X}$ and $\mathcal{Y}$ be two metric spaces. For a continuous map $T: \mathcal{X} \rightarrow \mathcal{Y}$, the push-forward operator $T_\sharp : \mathcal{M}(\mathcal{X})\rightarrow \mathcal{M}(\mathcal{Y})$ defined on the set $\mathcal{M}(\mathcal{X})$ of Radon measures on $\mathcal{X}$ is such that, for any $\alpha \in \mathcal{M}(\mathcal{X})$, the push-forward measure $\beta = T_\sharp \alpha \in \mathcal{M}(\mathcal{Y})$ of $\alpha$ satisfies for every continuous function $h:\mathcal{Y}\rightarrow \mathbb{R}$: $\int_{\mathcal{Y}} h(y)d\beta(y) = \int_\mathcal{X} h(T(x)) d\alpha(x)$.
\end{definition}
For a discrete measure $\alpha = \sum_i a_i \delta_{x_i}$, the push-forward measure is obtained by moving the positions of all the points in the support of the measure: $T_\sharp \alpha =\sum_i a_i \delta_{T(x_i)}$.

\begin{definition}[1-dimensional Wasserstein distance]
\label{def:appendix_w_1d}
If $X=\mathbb{R}$, then, the Wasserstein distance \ref{def:wasserstein} has the closed form expression:
\begin{equation}
     W_r(\mu, \nu) = \left( \int_0^1 \left| F^{-1}_\mu(t)-F^{-1}_\nu(t)\right|^r dt \right)^{\frac{1}{r}} 
\end{equation}
where $F_\mu(x)=\mu((-\infty, x]), x\in \mathbb{R}$ is the cumulative distribution function and $F_\mu^{-1}(t) = \inf \{x\in \mathbb{R} : F_\mu(x)\geq t\}, t\in [0,1]$ is the inverse cumulative distribution function (quantile function).
\end{definition}

\section{Details about the compared methods and their hyperparameters}

In this section we give more details on hyperparameters of reference methods for classification and regression tasks presented in the experimental section. We also explain the choice of the compared methods for regression. In particular, we emphasize the difficulties of 
applying other approaches to the large meshes of interest, hence we also consider coarsened meshes which make it possible to compare the different methods. Note that we used indefinite kernels for the kernels methods (SVM and Gaussian process regression), even if the existence of a reproducing kernel Hilbert space in this case is not true. Some work try to extend to the indefinite case in Krein spaces, but it remains a largely open question (see for example the work of \cite{liu2020analysis} for kernel ridge regression). We nevertheless test several graph kernels regardless of the positive definiteness as they are commonly used and give reasonable results in practice.

\subsection{Reference methods for classification of small graphs.}
In this section we give some additional details about the hyperparameter tuning for the SVM classification of small graphs and we recall the time complexities.

\emph{FGW}: The balance parameter $\alpha$ between structure and features is tuned in $\{0, 0.25, 0.5, 0.75, 1.0\}$.

\emph{RPW}: We use the same setting as in \cite{a_simple_way_to_learn_metrics_btw_attributed_graphs}. For the SGCN part, we use 10 iterations and a learning rate of $1e^{-2}$ (except for PROTEINS with 20 iterations, a learning rate of $1e^{-3}$ and ENZYMES with 20 iterations, a learning rate of $1e^{-4}$). The number of layers is optimized in $\{1,2,3\}$. The final layer dimension is fixed to 5. For the Restricted Wasserstein part, we take 50 uniformly sampled projections.

\emph{GH}: We use a RBF kernel to compare node attributes. The current implementation of Grakel (\cite{grakel}) uses a fixed lengthscale parameter of 1.

\emph{PK}: The results are given for a bin width $w=1e^{-2}$ and we optimize the number of iterations $\tau \in \{1,3,5,8,10,15,20\}$. We have also tested with varying bin width $w\in\{1e^{-5}, 1e^{-4}, \ldots, 1e^{-1}\}$ but we have found that $w=1e^{-2}$ gives the best scores.

We also remind time complexities of the kernels in Table \ref{tab:complexities}.

\begin{table*}[bp]
\setlength{\tabcolsep}{3pt}

\caption{Comparison of the complexities of the compared kernels to build a $N\times N$ Gram matrix where $n$ the average number of nodes, $\delta$ is the average degree of nodes, $\Delta$ is the maximum graph diamater.}
\label{tab:complexities} 
\centering
\begin{tabular}{ccccccc} 
 \toprule
 Kernel & Time Complexity & Quantities\\ 
 \midrule
 SWWL (ours) & $\mathcal{O}(NH\delta n$ + $NPn(\log(n)+H)+N^2PQ)$ & $P$: projections, $Q$: quantiles, $H$: WL iterations\\ 
 WWL & $\mathcal{O}(NH\delta n + HN^2n^3\log(n))$ & $H$: WL iterations \\ 
 FGW & $\mathcal{O}(Nn^2+N^2n^3)$ & - \\ 
 RPW & $\mathcal{O}(NHn^2+N^2Pn\log(n))$ & $H$: layers, $P$: last layer size\\ 
 PK & $\mathcal{O}(\tau(Nn+N^2n))$ & $\tau$: iterations \\ 
 GH & $\mathcal{O}(N(n^3+n^2\Delta^2)+N^2n^2)$ & - \\ 
 
 \bottomrule
\end{tabular}
\end{table*}
\vfill

\subsection{Reference methods for regression of large meshes.}
Two major problems arise when comparing to WWL. The time to calculate a Wasserstein distance between embeddings of size 30000 is approximately 80 seconds, so more than 400 days are needed to compute the pairwise distances of 1000 graphs. The second issue is memory usage. Allocating matrices of size 30000 $\times$ 30000 for the transport costs of the Wasserstein distance is only possible with a minimum amount of space and severely limits parallelization. We still include a comparison with WWL on coarsened datasets with 100 parallel jobs as it is the method on which SWWL is based. Remark that we do not use entropic regularization (\cite{sinkhorn}) that would reduce the complexity but has the cost of an extra hyperparameter to tune in an outer loop.

We do not include FGW and GH for time and memory issues. The custom loss of RPW designed for classification makes it not applicable for regression tasks. 

For the special case of PK, we face memory issues. This is mainly due to the transition matrices of all the graphs that are stored simultaneously in the implementation of \cite{grakel}. For the special case of \texttt{AirfRANS-CM}, we use a shared server running  Linux CentOS 7.9.2009, with 2 CPU (RAMIntel Xeon E5-2680 @ 2.50 GHz) using 24 cores, and a total of 378 GB of RAM. Remark that this computing infrastructure is not sufficient to build the PK kernel with \texttt{Rotor37}, \texttt{Tensile2d} and \texttt{AirfRANS}. 

We tested the PK with a bin width $w\in \{1e^{-5}, 1e^{-4}, \ldots, 1e^{-1} \}$ and a number of iterations $\tau \in \{1,3,5,7\}$. The results given in the paper correspond to $\tau=1$ and $w=1e^{-2}$ except for \texttt{Rotor37-CM} where $w=1e^{-3}$ as such parameters give the best test results. Note that these parameters should be finely tuned using a hyperparameter grid, thus drastically increasing the times in Table \ref{tab:regression_times}.

For $SWWL(\sqrt{n})$, the step of WL iterations is chosen to be respectively $T = 173, 30, 100, 30, 100, 100$ for
\texttt{Rotor37}, \texttt{Rotor37-CM}, \texttt{Tensile2d}, \texttt{Tensile2d-CM}, \texttt{AirfRANS},
\texttt{AirfRANS-CM}. These values correspond approximatively to the square root of the average number of nodes, except for AirFRANS where it is lower for computational reasons. They have therefore not been optimized.

The datasets \texttt{Rotor37},\texttt{Tensile2d}, and \texttt{AirfRANS} can be found according to the link given in the main paper. Please note that the datasets \texttt{Rotor37} that we used in the experiments and the one that is publicly available use different type of finite elements (quads/triangles) and the conversion from quads to triangles yields possibly different graphs. Scores may vary slightly with this public version.

\subsection{Robust Gaussian process regression}
Several strategies can be used to estimate the hyperparameters in Gaussian process regression. We use the robust Gaussian stochastic process emulation (RGaSP) (\cite{rgasp}) that estimates robustly the different lengthscale parameters. We first describe the method, we then give implementation details and we finally give the computation times (independent of the choice of the kernel).

\paragraph{Robust estimation of range parameters.}
We consider a Gaussian process $f \sim \mathcal{G}\mathcal{P}(m, k)$ where $m : \mathcal{X} \rightarrow \mathbb{R}$ is the mean function and $k: \mathcal{X}\times \mathcal{X} \rightarrow \mathbb{R}$ is the covariance function. We write $k(\cdot, \cdot) = \sigma^2 c(\cdot, \cdot)$ to separate the variance $\sigma^2$ and the correlation function $c$. In this section, we suppose that $m$ is a constant function determined by the parameter $\theta \in \mathbb{R}$. The correlation function has range parameters that we denote by $\bm{\gamma} = (\gamma_1, \ldots, \gamma_L)$ in an anisotropic setting (in our paper $\mathcal{X}$ is a set of graphs and $L=1$).

We assume that we have observations $y_i = f(\mathbf{x}_i)$ for $i\in \{1, \ldots, N\}$ of $f$ at input locations $\mathbf{X}=(\mathbf{x}_i)_{i=1}^N$. We consider the correlation matrix $\mathbf{R} \in \mathcal{M}_{N,N}(\mathbb{R})$ at input points  where $(\mathbf{R})_{ij} = c(\mathbf{x}_i, \mathbf{x}_j)$. Following the notations of section \ref{section:swwl_for_gp}, the train Gram matrix is thus $\mathbf{K} = \sigma^2 \mathbf{R}$. We omit the white noise that only changes the train covariance matrices.

The common way to deal with the mean and variance parameters is to use their closed form in a Bayesian setting. These parameters are thus assigned to the objective prior $\pi(\theta,\sigma^2) \propto \frac{1}{\sigma^2}$. This prior is then used to marginalize out the mean and variance parameters in the likelihood function and lead to the marginal likelihood
\begin{equation}
\label{eq:marginal_likelihood}
    \mathcal{L}(\bm{\gamma} | \mathbf{X}, \mathbf{y}) \propto |\mathbf{R}|^{-\tfrac{1}{2}} |\mathbf{h}^T \mathbf{R}^{-1} \mathbf{h}|^{-\tfrac{1}{2}} 
(S^2)^{-\frac{n-1}{2}}  ,
\end{equation}
where $\mathbf{h}$ is the $N\times 1$ column vector where all coefficients are equal to 1, $S^2 = \mathbf{y}^T \mathbf{R}^{-1} \left[ \mathbf{I_N} - \mathbf{h} (\mathbf{h}^T \mathbf{R}^{-1} \mathbf{h})^{-1} \mathbf{h}^T \mathbf{R}^{-1}  \right] \mathbf{y}$.
However, the maximum marginal likelihood estimators obtained by maximizing \eqref{eq:marginal_likelihood} are not robust in the sense that they can lead to covariance matrices close to the identity or to a matrix made up of one, which give rise to numerical issues.

In the robust Gaussian process framework, the marginal likelihood is augmented by a prior for the range parameters
$\pi(\theta, \sigma^2, \bm{\gamma}) \propto \frac{\pi(\mathbf{\gamma)}}{\sigma^2}$. We use 
the prior $\pi^{JR}(\bm{\gamma})  = \left( \sum_{l=1}^L C_l \gamma_l^{-1} \right)^a \exp\left( -b \sum_{l=1}^{L} C_l \gamma_l^{-1} \right) $, 
$a=0.2$, $b=N^{-1/L}(a+L)$, and $C_l$ equal to the mean of $|x_{il}-x_{il}|$ for $1\leq i,j\leq N, i\neq j$.
Remark that in the case where $x_{il}$ is not a scalar but rather a structured object, we replace the absolute value by the user-defined distance or pseudo-distance that is substituted in the kernel (in our paper we use the estimated sliced Wasserstein distance).

The range parameters are estimated by maximization of the marginal posterior distribution
\begin{equation}
    (\hat{\gamma_1}, \ldots, \hat{\gamma_L}) = \argmax\limits_{\gamma_1, \ldots, \gamma_L}\{ \mathcal{L}(\gamma |\mathbf{X},\mathbf{y}) \pi^{JR}(\bm{\gamma}) \}.
\end{equation}

With the use of the estimated range parameters, the predictive distribution of unknown targets $\mathbf{f}_*:=(f(\mathbf{x}^*_i))_{i=1}^{_{}N^*}$ at $N^*$ new locations $\mathbf{X}^*=(\mathbf{x}^*_i)_{i=1}^{N^*}$ given $\mathbf{X}$, $\mathbf{y}$ and $\bm{\gamma}$ is a Student's $t$-distribution with $N-1$ degrees of freedom
\begin{equation}
    \mathbf{f}_* | \mathbf{X}, \mathbf{y}, \mathbf{X}^*, \gamma \sim t(\mathbf{\bar{m}}, \hat{\sigma}^2 \mathbf{\bar{C}}, N-1) ,
\end{equation}
where
\begin{align}
\mathbf{\bar{m}}&= \mathbf{h}_* \hat{\theta} + \mathbf{R}_* \mathbf{R}^{-1} (\mathbf{y}-\mathbf{h}\hat{\theta}),\\
\hat{\sigma}^2 &= (N-1)^{-1} (\mathbf{y}-\mathbf{h}\hat{\theta})^T \mathbf{R}^{-1} (\mathbf{y}-\mathbf{h}\hat{\theta}),\\
\mathbf{\bar{C}} &= \mathbf{R}_{**} - \mathbf{R}_{*}\mathbf{R}^{-1} \mathbf{R}_{*}^T + (\mathbf{h}_* - \mathbf{h}^T \mathbf{R}^{-1} \mathbf{R}_{*}^T)^T (\mathbf{h}^T \mathbf{R}^{-1} \mathbf{h})^{-1} (\mathbf{h}_*-\mathbf{h}^T \mathbf{R}^{-1} \mathbf{R}^T_{*}),
\end{align}
with $\mathbf{h}_*$ being the $N^* \times 1$ vector composed of ones, $\hat{\theta}=(\mathbf{h}^T \mathbf{R}^{-1} \mathbf{h})^{-1} \mathbf{h}^T \mathbf{R}^{-1} \mathbf{y}$ being the least squares estimator for $\theta$, and $\mathbf{R}_{*}$, $\mathbf{R}_{**}$ being the covariance matrices evaluated between train inputs and test/train inputs, respectively.



\paragraph{Computation times.} 
Once distance matrices are precomputed, for \texttt{Rotor37} (1000 train inputs), the training takes 220 seconds;  for \texttt{Tensile2d} (500 train inputs), the training takes 70 seconds; for \texttt{AirfRANS} (800 train inputs), the training takes 135 seconds. This time is similar for coarsened meshes as the shape of distance matrices is the same.



\section{Additional advantages from GP regression}

We finally discuss some benefits coming from GP regression that prove useful in practice for physics-based learning problems.

\subsection{Anisotropic variant of SWWL}

The difficulty to choose the number of WL iterations has led us to define an anisotropic variant of SWWL. The idea is to leverage the information of each iteration separately by using $H+1$ tensorized SWWL kernels.

\begin{definition}
    Let $H$ be a number of WL iterations and $G\in \mathcal{G}$ be a graph with its continuous WL embedding $\mathbf{E}^G = ( \mathbf{E}^{G}_u)_{u\in V}$, $ \mathbf{E}^{G}_u   \in \mathbb{R}^{(H+1)d}$. We denote by $\mathbf{E}^{G,(h)}  = ( \mathbf{E}^{G,(h)}_u)_{u\in V} $,  $\mathbf{E}^{G,(h)}_u\in \mathbb{R}^{d}$, its $h$-th continuous WL iteration and by $\mu^{(h)}_G = |V|^{-1}\sum_{ u\in V} \delta_{(\mathbf{E}^{G,(h)})_u}$ the associated empirical measure for $h\in \{0,\ldots, H\}$.

    The ASWWL kernel is defined as follows:\\
    \begin{equation}
    k_{\mathrm{ASWWL}}(G,G') = \prod_{h=0}^{H} \exp \left(-\gamma_h \widehat{SW}^2_{2,P,Q}(\mu^{(h)}_G, \mu^{(h)}_{G'}) \right) ,
    \end{equation} 
    for $\gamma_0 >0, \gamma_1>0, \ldots, \gamma_H > 0$ the precision parameters. 
    
\end{definition}

The ASWWL kernel is positive definite as a tensor product of positive defined kernels.
The time complexity of ASWWL is:
$O(NH\delta n + NPHn\log(n) +N^2HPQ)$ as we now need to project and sort the data for all iterations and build $H+1$ distance matrices.

The advantage of this approach is to better control the weights given at each iteration to make an automatic selection of interesting iterations. The biggest downside with this approach is the optimization of the precision parameters $\gamma_0, \ldots, \gamma_H$ that increases the cost of learning with a Gaussian process regression (or other kernel regression method) and thus limits the number of iterations to keep.\\

\subsection{Application of uncertainty quantification: characteristic curves}
We present an example of the use of GP regression to obtain prediction uncertainties for a practical computational fluid dynamics scenario.

We focus here on the dataset \texttt{Rotor37} where inputs are triplets composed of the finite element mesh of a 3D compressor blade, the (scalar) rotational speed and the (scalar) input pressure. During the design of such 3D compressor blades, characteristic curves are often used to visualize the evolution of several physical quantities in different operating regimes. Such characteristic curves allow the engineer to study the influence of geometric variations, which are relevant to determining an optimum geometric shape and operating parameters. When considering a new blade, the engineer does not necessarily know if it belongs to the training distribution, so he needs to know the uncertainties to assign a level of confidence to the prediction.

\begin{figure}[h]
\begin{center}
\includegraphics[scale=0.4]{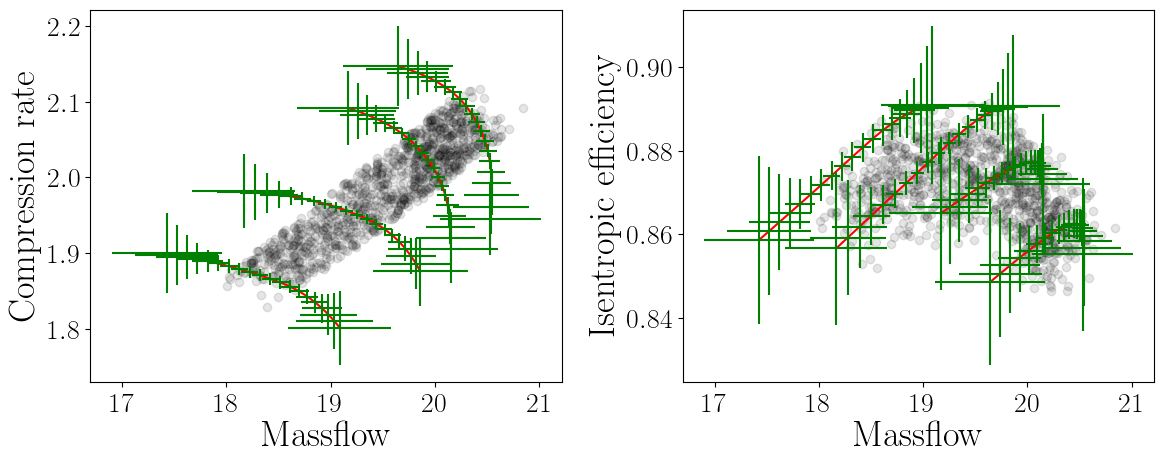}
\end{center}
\caption{Graphs of the predicted compression rate and isentropic efficiency with respect to the
massflow for a test mesh, 4 input rotations, and 20 values of input pressures that go beyond the range
of the train and test datasets. The green lines correspond to 95\% confidence intervals. Black dots correspond to train outputs.}
\label{fig:charac}
\end{figure}

Figure \ref{fig:charac} shows four curves corresponding to four different input rotations and twenty input pressures that go beyond the ranges of the train input pressures for a new test blade. We consider two extra (scalar) output quantities of interest, namely massflow and compression rate (not mentioned in the paper) that characterize the engine performance. Confidence intervals for the three (scalar) outputs are represented. We observe that the uncertainties are reduced near the training domain compared to the uncertainties obtained when the model is applied outside the support of the training distribution. The Gaussian process therefore gives coherent uncertainties when using the SWWL kernel.

\vfill

\end{document}